\documentclass{article}

\usepackage[accepted]{icml2026} 
\usepackage{times}
\usepackage{latexsym}
\usepackage{amsmath}
\usepackage{amssymb}
\usepackage{amsthm}
\usepackage{graphicx}
\usepackage{booktabs}
\usepackage{algorithm}
\usepackage{algorithmic}
\usepackage{hyperref}
\usepackage{url}
\usepackage{multirow}
\usepackage{xcolor}
\usepackage{subcaption}
\usepackage{wrapfig}
\usepackage{colortbl}
\usepackage{tabularx}
\usepackage{tcolorbox}

\newcommand{\method}{Agentic-VLA}

\newcommand{\eg}{\textit{e.g.}}

\definecolor{lightgray}{gray}{0.92}

\newcolumntype{Y}{>{\centering\arraybackslash}X}

\icmltitlerunning{Agentic-VLA: Efficient Online Adaptation for Vision-Language-Action Models}

\begin{document}

\twocolumn[
  \icmltitle{Agentic-VLA: Efficient Online Adaptation for Vision-Language-Action Models}

  \icmlsetsymbol{equal}{*}

\begin{icmlauthorlist}
\icmlauthor{Ruofan Jin}{scinetics}
\icmlauthor{Zaixi Zhang}{scinetics}
\end{icmlauthorlist}

\icmlaffiliation{scinetics}{Scinetics}

\icmlcorrespondingauthor{Zaixi Zhang}{zaixizhang8@gmail.com}

\icmlkeywords{Machine Learning, ICML}

  \vskip 0.3in
]

\printAffiliationsAndNotice{}  

\begin{abstract}
Vision-Language-Action (VLA) models have emerged as a promising paradigm for robotic manipulation by leveraging pre-trained vision-language representations. However, current VLA training methods suffer from two critical limitations: \textit{poor generalization} to novel environments and \textit{low training efficiency} requiring extensive demonstrations. We introduce \method{}, an agentic training framework that enables VLAs to efficiently adapt online through three key innovations: (1) \textbf{Adaptive Reward Synthesis}, which dynamically generates and adjusts reward functions based on the VLA's current capabilities and task complexity, decomposing complex tasks into learnable sub-goals for curriculum learning; (2) \textbf{Language-Guided Exploration}, where a critic model provides structured guidance for systematic exploration rather than random sampling; and (3) \textbf{Experience Memory}, which stores and retrieves task-relevant policy weights for warm-starting adaptation to similar tasks. We evaluate \method{} on the LIBERO benchmark, achieving substantial improvements: +12.3\% on long-horizon tasks, +28.5\% in 1-shot learning, and enabling cross-task transfer from 0\% to 31.2\% without task-specific demonstrations. Our framework also demonstrates 2.4$\times$ faster convergence compared to existing online adaptation methods. Beyond LIBERO, \method{} retains its advantage on the dual-arm RoboTwin~2.0 benchmark, including under its randomized Hard setting. These results establish \method{} as a significant step toward truly adaptive VLA systems capable of continuous learning in deployment.
\end{abstract}

\section{Introduction}
\label{sec:intro}

The pursuit of general-purpose robots capable of performing diverse manipulation tasks has been a long-standing goal in robotics and artificial intelligence. Recent advances in Vision-Language-Action (VLA) models~\citep{brohan2022rt, kim2024openvla, black2024pi0} have demonstrated remarkable progress by leveraging the semantic understanding of large language models to interpret task instructions and generate corresponding actions. However, despite their impressive capabilities, current VLA models face fundamental limitations that hinder their deployment in real-world scenarios.

\textbf{The Generalization Challenge.} VLA models trained through Supervised Fine-Tuning (SFT) on expert demonstrations exhibit brittle generalization. They tend to memorize specific trajectories rather than learning the underlying task semantics, leading to failure when deployment conditions deviate even slightly from training distributions. A single misstep during execution often cascades into complete task failure, as these models lack mechanisms for error recovery or adaptive replanning.

\textbf{The Efficiency Challenge.} Current VLA training paradigms require hundreds of demonstrations per task, with data collection costs scaling linearly with the number of tasks. This bottleneck makes it impractical to scale VLAs to the diverse range of tasks required for general-purpose robotics.

Recent work has begun exploring online adaptation for VLAs~\citep{li2025simplevla, guo2025improving}, demonstrating that reinforcement learning can improve performance beyond pure imitation. However, these approaches face significant challenges: (1) reliance on noisy reward signals that can mislead learning, (2) inefficient random exploration that wastes computational resources, and (3) inability to transfer knowledge across related tasks.

In this work, we propose \method{}, an \textit{agentic} training framework that fundamentally reimagines how VLAs learn and adapt. Our key insight is that the learning process itself should be \textit{intelligent}: rather than treating reward design, exploration, and knowledge transfer as fixed procedures, we employ learned agents that actively orchestrate these components based on the current learning state.

As illustrated in Figure~\ref{fig:teaser}, \method{} introduces three synergistic innovations:

\begin{itemize}
    \item \textbf{Adaptive Reward Synthesis}: Instead of fixed reward functions, we employ a reward agent that analyzes the VLA's current capabilities and dynamically generates appropriate reward signals. For complex tasks, the agent automatically decomposes them into sub-goals, creating a natural curriculum that guides learning from simple to complex behaviors.
    
    \item \textbf{Language-Guided Exploration}: We replace random exploration with structured, language-guided exploration. A critic agent observes the VLA's behavior and provides natural language suggestions (\eg, ``try approaching the object from the left side''), enabling systematic discovery of effective strategies rather than blind trial-and-error.
    
    \item \textbf{Experience Memory}: We maintain a memory bank of policy weights indexed by task embeddings. When encountering a new task, our framework retrieves and adapts weights from similar previously learned tasks, enabling rapid warm-start adaptation.
\end{itemize}

We evaluate \method{} on the LIBERO benchmark~\citep{liu2023libero} and the dual-arm RoboTwin~2.0 benchmark~\citep{chen2025robotwin}, demonstrating substantial improvements across all metrics. Our contributions include:

\begin{enumerate}
    \item We propose \method{}, the first agentic framework for VLA online adaptation that integrates adaptive reward synthesis, language-guided exploration, and experience memory into a unified system.
    
    \item We introduce a novel adaptive reward synthesis mechanism that automatically decomposes complex tasks and adjusts reward granularity based on learning progress.
    
    \item We demonstrate that language-guided exploration significantly improves sample efficiency by enabling structured, goal-directed exploration.
    
    \item We show that experience memory enables effective cross-task transfer, achieving non-trivial success rates on tasks without any task-specific demonstrations.
    
    \item Extensive experiments demonstrate state-of-the-art performance: +12.3\% on long-horizon tasks, +28.5\% in 1-shot learning, 2.4$\times$ faster convergence, and cross-task transfer from 0\% to 31.2\%. Controlled comparisons against alternative curricula and exploration bonuses, multi-seed results, and evaluation on RoboTwin~2.0 further confirm that the gains stem from our specific design rather than from generic components or seed variance.
\end{enumerate}

\section{Related Work}
\label{sec:related}

\subsection{Vision-Language-Action Models}

Vision-Language-Action models represent a paradigm shift in robot learning by leveraging pre-trained vision-language representations for action generation. RT-1~\citep{brohan2022rt} pioneered this direction by training transformers on large-scale robotic data. Subsequent works have explored different architectural choices: OpenVLA~\citep{kim2024openvla} provides an open-source implementation with standardized evaluation, OpenVLA-OFT~\citep{kim2025fine} introduces parallel decoding and action chunking for improved efficiency, and $\pi_0$~\citep{black2024pi0} proposes a flow-based architecture for continuous action generation.

Despite impressive results, these models are trained exclusively through imitation learning, inheriting its fundamental limitations: dependence on expert demonstrations, inability to improve through experience, and poor generalization beyond training distributions. Our work addresses these limitations by enabling online adaptation.

\subsection{Reinforcement Learning for VLAs}

Recent work has begun exploring RL fine-tuning for VLA models. VLA-RL~\citep{lu2025vla} introduces trajectory-level RL formulation, SimpleVLA-RL~\citep{li2025simplevla} and $\pi_{RL}$~\citep{chen2025pirl} explore different policy optimization methods, and RL4VLA~\citep{liu2025can} provides systematic studies of RL's impact across different dimensions. EVOLVE-VLA~\citep{bai2025evolvevla} proposes test-time training with learned progress estimators.

However, all these approaches assume access to oracle reward signals or rely on noisy learned rewards without mechanisms to handle their imperfections. Our work advances this line by introducing adaptive reward synthesis that adjusts to the VLA's capabilities, language-guided exploration for efficient learning, and experience memory for cross-task transfer.

\subsection{Curriculum Learning in Robotics}

Curriculum learning has been widely explored to improve sample efficiency in robot learning. In the context of VLA models, recent work on RL fine-tuning implicitly incorporates curriculum ideas through progressive training strategies. For instance, EVOLVE-VLA~\citep{bai2025evolvevla} introduces progressive horizon extension that gradually increases task complexity during training. Similarly, works on vision-language reward models~\citep{ma2023liv, sontakke2023roboclip} enable learning from sparse demonstrations by providing shaped feedback signals. Our adaptive reward synthesis mechanism extends these ideas by dynamically adjusting reward functions based on the VLA's evolving capabilities, creating a fine-grained curriculum at the sub-goal level rather than at the task or horizon level. As we show empirically (Section~\ref{sec:control_ablations}), this capability-aware weighting outperforms uniform, fixed-schedule, and learning-progress-based curricula under a matched training budget.

\subsection{Memory and Transfer in Robot Learning}

Transfer learning in robotic manipulation has primarily focused on leveraging pre-trained vision-language representations. OpenVLA~\citep{kim2024openvla} and Octo~\citep{team2024octo} demonstrate that large-scale pre-training enables zero-shot transfer to new robot embodiments. Recent work on RL fine-tuning for VLAs~\citep{li2025simplevla, lu2025vla} shows that online adaptation can further improve generalization. However, these approaches learn each task independently without explicit mechanisms for cross-task knowledge sharing. Our experience memory mechanism addresses this gap by maintaining a bank of adapted policy weights indexed by task semantics, enabling warm-start initialization that accelerates learning on related tasks.

\begin{figure*}[t]
    \centering
    \includegraphics[width=1.0\linewidth]{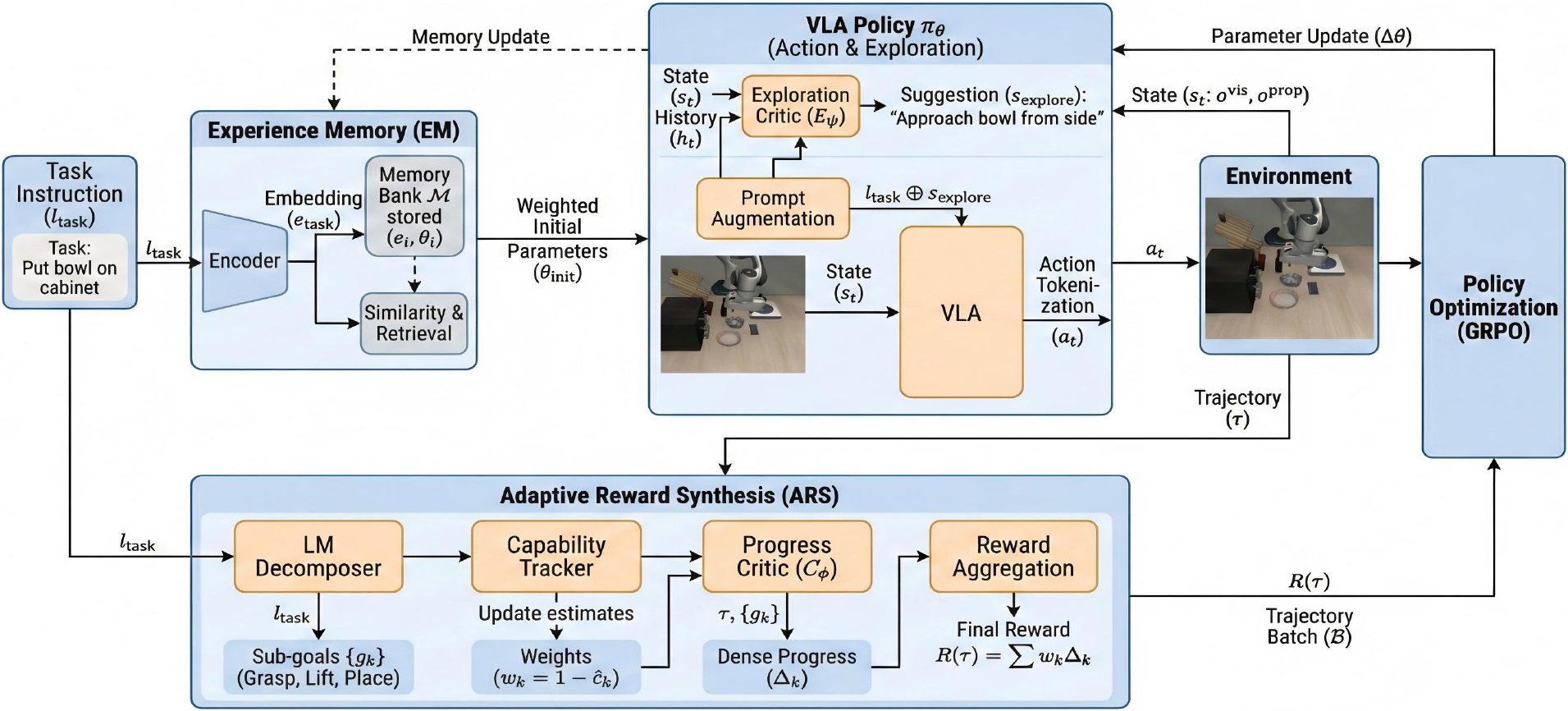}
    \vspace{-10pt}
    \caption{
        \textbf{Framework overview.} 
        The adaptation loop begins with \textbf{Experience Memory} retrieving relevant parameters for warm initialization. 
        During interaction, the VLA receives structured guidance from \textbf{Language-Guided Exploration}, which generates prompt-based hints to aid diverse behavior discovery. 
        Trajectories are evaluated by \textbf{Adaptive Reward Synthesis}, which dynamically weighs sub-goals based on the agent's real-time capability estimates. 
        This capability-aware mechanism naturally forms an auto-curriculum, providing stable dense feedback for \textbf{GRPO optimization}.
    }
    \label{fig:teaser}
    \vspace{-10pt}
\end{figure*}

\section{Method}
\label{sec:method}

\subsection{Problem Formulation}

We formulate robotic manipulation as a Markov Decision Process $\mathcal{M} = (\mathcal{S}, \mathcal{A}, P, R, \gamma)$, where $\mathcal{S}$ is the state space, $\mathcal{A}$ is the action space, $P$ represents transition dynamics, $R$ is the reward function, and $\gamma \in [0, 1)$ is the discount factor. At timestep $t$, the state $s_t = (o_t^{\text{vis}}, o_t^{\text{prop}}, l_{\text{task}})$ consists of visual observation $o_t^{\text{vis}}$, proprioceptive state $o_t^{\text{prop}}$, and task instruction $l_{\text{task}}$.

A VLA policy $\pi_\theta: \mathcal{S} \rightarrow \Delta(\mathcal{A})$ maps states to action distributions. Following modern VLA architectures, we adopt action tokenization where continuous actions are discretized into tokens, with the policy autoregressively generating action sequences.

\textbf{Key Challenge.} Traditional RL assumes access to a well-defined reward function $R$. At deployment time, however, such oracle rewards are unavailable. Our goal is to enable effective online adaptation using only self-generated feedback signals.

\subsection{Framework Overview}

Figure~\ref{fig:teaser} presents an overview of \method{}. Our framework consists of three core components that work synergistically:

\begin{enumerate}
    \item \textbf{Adaptive Reward Synthesis (ARS)}: Analyzes the current VLA capabilities and task requirements to dynamically generate and adjust reward functions.
    
    \item \textbf{Language-Guided Exploration (LGE)}: Provides structured exploration guidance through natural language suggestions.
    
    \item \textbf{Experience Memory (EM)}: Stores and retrieves task-relevant policy weights for warm-starting.
\end{enumerate}

The training loop proceeds as follows: Given a new task, Experience Memory retrieves relevant prior weights for initialization. The VLA then interacts with the environment, with Language-Guided Exploration providing suggestions for behavior. Adaptive Reward Synthesis evaluates trajectories and generates appropriate rewards. The policy is updated via GRPO~\citep{shao2024deepseekmath}, and successful adaptations are stored back in Experience Memory.

\subsection{Adaptive Reward Synthesis}
\label{sec:ars}

A fundamental challenge in VLA online adaptation is the absence of oracle reward signals. While learned progress estimators~\citep{zhai2025vlac} provide a proxy, they are inherently noisy and may not align with the VLA's current learning needs. We address this through Adaptive Reward Synthesis (ARS), which dynamically generates reward functions tailored to the VLA's capabilities.

\subsubsection{Task Decomposition}

For complex, long-horizon tasks, directly optimizing for final success is inefficient due to sparse rewards. ARS automatically decomposes tasks into sub-goals based on the task instruction.

Given task instruction $l_{\text{task}}$, we use a language model to generate a sequence of sub-goals:
\begin{equation}
    \mathcal{G} = \{g_1, g_2, \ldots, g_K\} = \text{LM}_{\text{decompose}}(l_{\text{task}})
\end{equation}
where each $g_k$ represents an intermediate milestone toward task completion. For example, the task ``turn on the stove and put the moka pot on it'' is decomposed into: $g_1$: ``locate and approach the stove'', $g_2$: ``turn on the stove'', $g_3$: ``locate the moka pot'', $g_4$: ``grasp the moka pot'', $g_5$: ``place the pot on the stove''.

\subsubsection{Capability-Aware Reward Adjustment}

Rather than using fixed reward magnitudes, ARS dynamically adjusts sub-goal rewards based on the VLA's current proficiency. We maintain a capability estimate $\hat{c}_k \in [0, 1]$ for each sub-goal $g_k$, computed as an exponential moving average of recent success rates:
\begin{equation}
    \hat{c}_k^{(t+1)} = \alpha \cdot \hat{c}_k^{(t)} + (1 - \alpha) \cdot \mathbb{I}[\text{success at } g_k]
\end{equation}
where $\alpha \in [0, 1]$ is a smoothing coefficient that controls the adaptation speed.

The reward weight for each sub-goal is then computed as:
\begin{equation}
    w_k = 1 - \hat{c}_k
\end{equation}
This simple formulation implements a natural curriculum: sub-goals where the VLA struggles ($\hat{c}_k \approx 0$) receive full weight ($w_k \approx 1$) to encourage focused learning, while mastered sub-goals ($\hat{c}_k \approx 1$) receive diminished weight ($w_k \approx 0$) to shift attention toward remaining challenges.

\subsubsection{Progress-Based Dense Rewards}

For each sub-goal, we employ a learned critic $C_\phi$ to provide dense progress estimates. Inspired by accumulative progress estimation~\citep{bai2025evolvevla}, the reward for a trajectory $\tau$ is computed as:
\begin{equation}
    R(\tau) = \sum_{k=1}^{K} w_k \cdot \Delta_k(\tau)
\end{equation}
where $\Delta_k(\tau) = C_\phi(o_{\text{start}}^k, o_{\text{end}}^k, g_k)$ measures progress on sub-goal $g_k$ with critic model VLAC \citep{zhai2025vlac}, and $w_k$ are the capability-adjusted weights defined above. This formulation unifies the dense progress signal with the adaptive weighting mechanism, providing informative feedback that automatically focuses on the VLA's current learning frontier.

\subsection{Language-Guided Exploration}
\label{sec:lge}

Standard RL exploration relies on random sampling from action distributions, which is inefficient for high-dimensional VLA action spaces. We introduce Language-Guided Exploration (LGE), which leverages a pre-trained Vision-Language Model (VLM) to provide structured exploration guidance through in-context reasoning.

\subsubsection{VLM-Based Exploration Critic}

Unlike prior approaches that require training specialized critic models, we leverage the zero-shot reasoning capabilities of pre-trained VLMs to analyze the robot's current situation and generate actionable suggestions. Given the current visual observation, task instruction, and recent trajectory context, we prompt a VLM to identify potential issues and suggest improvements:
\begin{equation}
    s_{\text{explore}} = \text{VLM}(o_t^{\text{vis}}, l_{\text{task}}, l_{\text{prompt}})
\end{equation}
where $l_{\text{prompt}}$ is a carefully designed prompt template that instructs the VLM to act as a manipulation expert, and $s_{\text{explore}}$ is the generated language suggestion.

This approach enables the critic to leverage the VLM's pre-trained knowledge of spatial reasoning, object affordances, and manipulation strategies without any task-specific training. Example generated suggestions include:
\begin{itemize}
    \item ``The gripper is approaching from an occluded angle; try positioning from the left side''
    \item ``The grasp point is too close to the object edge; target the center for stability''
    \item ``Previous attempts overshot the target; reduce movement magnitude''
\end{itemize}

\subsubsection{Suggestion-Conditioned Action Generation}

The exploration suggestions are incorporated into the VLA's action generation through prompt augmentation:
\begin{equation}
    a_t \sim \pi_\theta(a | s_t, l_{\text{task}} \oplus s_{\text{explore}})
\end{equation}
where $\oplus$ denotes concatenation of the task instruction with the exploration suggestion.

This mechanism biases the action distribution toward suggested behaviors while maintaining stochasticity for continued exploration. Importantly, the suggestions are generated only during the exploration phase; during evaluation, the VLA operates with the original task instruction alone.

\subsubsection{Adaptive Suggestion Frequency}

We dynamically adjust the frequency of exploration suggestions based on learning progress. Early in training, suggestions are provided frequently to guide initial behavior discovery. As the policy improves, suggestion frequency decreases to allow the policy to develop autonomous competence:
\begin{equation}
    p_{\text{suggest}}(t) = p_{\text{max}} \cdot \exp(-\lambda \cdot \bar{R}^{(t)})
\end{equation}
where $\bar{R}^{(t)}$ is the smoothed recent reward and $\lambda$ controls decay rate.

\subsection{Experience Memory}
\label{sec:em}

Learning each task from scratch ignores potentially transferable knowledge from related tasks. We introduce Experience Memory (EM), a mechanism for storing and retrieving task-relevant policy parameters.

\subsubsection{Memory Structure}

The memory bank $\mathcal{M}$ stores tuples $(e_i, \theta_i, m_i)$ where:
\begin{itemize}
    \item $e_i \in \mathbb{R}^d$ is the task embedding computed from the instruction
    \item $\theta_i$ are the adapted policy parameters
    \item $m_i$ contains metadata (success rate, training iterations, task complexity)
\end{itemize}

Task embeddings are computed using a frozen language encoder:
\begin{equation}
    e = \text{Encoder}(l_{\text{task}})
\end{equation}

\subsubsection{Warm-Start Retrieval}

When encountering a new task with instruction $l_{\text{new}}$, we compute its embedding $e_{\text{new}}$ and retrieve the $k$ most similar stored tasks:
\begin{equation}
    \mathcal{N}_k = \text{top-}k_{(e_i, \theta_i, m_i) \in \mathcal{M}} \cos(e_{\text{new}}, e_i)
\end{equation}

The initial policy parameters are computed as a weighted combination:
\begin{equation}
    \theta_{\text{init}} = \sum_{(e_j, \theta_j, m_j) \in \mathcal{N}_k} \frac{\exp(\cos(e_{\text{new}}, e_j) / \tau)}{\sum_{j'} \exp(\cos(e_{\text{new}}, e_{j'}) / \tau)} \cdot \theta_j
\end{equation}
where $\tau$ is a temperature parameter.
This soft retrieval mechanism enables graceful interpolation between related tasks, providing better initialization than using any single prior task.

\paragraph{Why weight-space interpolation is well-behaved here.} Averaging weights of independently trained networks is generally fragile, since unrelated solutions may lie in incompatible loss basins. Our use of EM is substantially more constrained. First, all retrieved policies share the same architecture and originate from the \emph{same} pretrained OpenVLA-OFT initialization, and the subsequent task-specific adaptations remain relatively local; the interpolation is therefore a warm-start heuristic among nearby task-specialized variants rather than unrestricted model merging. Second, retrieval is sharply peaked: we keep only the top-$k$ nearest tasks and weight them with a low temperature ($\tau{=}0.1$), so the initialization is dominated by the most semantically similar entries rather than a broad average over the bank. Third, the interpolated parameters are used only as an \emph{initialization} for subsequent online adaptation, which can correct an imperfect warm start. Empirically (Section~\ref{sec:ablations}), this initialization helps rather than harms.

\subsubsection{Memory Update}

After successful adaptation to a new task, we update the memory bank:
\begin{equation}
    \mathcal{M} \leftarrow \mathcal{M} \cup \{(e_{\text{new}}, \theta_{\text{adapted}}, m_{\text{new}})\}
\end{equation}

To prevent unbounded memory growth, we employ a capacity management strategy that prioritizes diverse, high-quality entries based on task embedding coverage and adaptation success rate.

\subsection{Training Algorithm}
\label{sec:training}

Algorithm~\ref{alg:training} presents the complete \method{} training procedure. The framework integrates all three components in a unified online adaptation loop.

\begin{algorithm}[t]
\caption{Agentic-VLA Training}
\label{alg:training}
\begin{algorithmic}[1]
\REQUIRE Task instruction $l_{\text{task}}$, base VLA $\pi_{\theta_0}$, memory bank $\mathcal{M}$
\ENSURE Adapted policy $\pi_\theta$

\STATE \textbf{// Experience Memory: Warm Start}
\STATE $e_{\text{task}} \leftarrow \text{Encoder}(l_{\text{task}})$
\STATE $\theta \leftarrow \text{WarmStartRetrieval}(\mathcal{M}, e_{\text{task}})$

\STATE \textbf{// Adaptive Reward Synthesis: Task Decomposition}
\STATE $\mathcal{G} \leftarrow \text{LM}_{\text{decompose}}(l_{\text{task}})$
\STATE Initialize capability estimates $\{\hat{c}_k\}_{k=1}^{|\mathcal{G}|}$

\FOR{iteration $= 1$ to $N_{\text{iter}}$}
    \STATE \textbf{// Rollout Generation with LGE}
    \STATE $\mathcal{B} \leftarrow \emptyset$
    \FOR{$i = 1$ to $N_{\text{batch}}$}
        \STATE $\tau_i \leftarrow \text{GenerateRollout}(\pi_\theta, E_\psi)$
        \STATE $\mathcal{B} \leftarrow \mathcal{B} \cup \{\tau_i\}$
    \ENDFOR
    
    \STATE \textbf{// Adaptive Reward Computation}
    \FOR{$\tau_i \in \mathcal{B}$}
        \STATE $R_i \leftarrow \text{ComputeAdaptiveReward}(\tau_i, \mathcal{G}, \{\hat{c}_k\})$
    \ENDFOR
    
    \STATE \textbf{// Policy Update (GRPO)}
    \STATE $\theta \leftarrow \text{GRPO}(\theta, \mathcal{B}, \{R_i\})$
    
    \STATE \textbf{// Update Capability Estimates}
    \STATE Update $\{\hat{c}_k\}$ based on sub-goal successes in $\mathcal{B}$
\ENDFOR

\STATE \textbf{// Memory Update}
\STATE $\mathcal{M} \leftarrow \mathcal{M} \cup \{(e_{\text{task}}, \theta, m)\}$

\STATE \textbf{Return} $\pi_\theta$
\end{algorithmic}
\end{algorithm}

\section{Experiments}
\label{sec:experiments}

\subsection{Experimental Setup}

\textbf{Benchmark.} We primarily use the LIBERO benchmark~\citep{liu2023libero}, which consists of four task suites: LIBERO-Spatial/Object/Goal/Long. Each suite contains 50 expert demonstrations per task. To test generality beyond LIBERO, we additionally evaluate on the dual-arm RoboTwin~2.0 benchmark~\citep{chen2025robotwin} (Section~\ref{sec:robotwin}).

\textbf{Metrics.} We report two metrics: (1) \textbf{Success Rate (SR)}: the percentage of trials where the task is fully completed, evaluated over 50 trials per task; and (2) \textbf{Progress}: the average normalized task completion measured by the VLAC critic, ranging from 0 (no progress) to 1 (task completed). Progress captures partial task completion even when full success is not achieved. Unless otherwise noted, LIBERO numbers are reported as mean $\pm$ standard deviation over 5 independent seeds.

\textbf{Base Model.} We apply \method{} to OpenVLA-OFT~\citep{kim2025fine}, a state-of-the-art autoregressive VLA model with action chunking and parallel decoding. We use discrete action tokens compatible with RL.

\textbf{Reward Model.} For progress estimation in ARS, we employ VLAC~\citep{zhai2025vlac}, a foundation critic model pre-trained on large-scale robotic manipulation data.

\textbf{Baselines.} We compare against:
\begin{itemize}
    \item \textbf{SFT methods}: Octo~\citep{team2024octo}, OpenVLA~\citep{kim2024openvla}, OpenVLA-OFT~\citep{kim2025fine}, $\pi_0$~\citep{black2024pi0}
    \item \textbf{RL methods}: VLA-RL~\citep{lu2025vla}, SimpleVLA-RL~\citep{li2025simplevla}, EVOLVE-VLA~\citep{bai2025evolvevla}
\end{itemize}

\textbf{Implementation Details.} We use GRPO~\citep{shao2024deepseekmath} for policy optimization with learning rate $1 \times 10^{-5}$, batch size 32, and group size 8. For ARS, we use Llama-3-8B for task decomposition. The exploration critic is based on a fine-tuned CLIP model. Memory bank capacity is set to 100 entries with $k=3$ for retrieval. All experiments use 4 NVIDIA A100 GPUs.

\subsection{Main Results}
\label{sec:main_results}

\begin{table}[t]
\centering
\caption{Main results on LIBERO benchmark, reported as mean $\pm$ std over 5 seeds for the methods we run. $\dagger$: reproduced with our settings. Best results in \textbf{bold}, second best \underline{underlined}.}
\label{tab:main_results}
\resizebox{\columnwidth}{!}{%
\begin{tabular}{lccccc}
\toprule
\textbf{Model} & \textbf{Spatial} & \textbf{Object} & \textbf{Goal} & \textbf{Long} & \textbf{Avg} \\
\midrule
\multicolumn{6}{l}{\textit{SFT Methods}} \\
Octo & 78.9 & 85.7 & 84.6 & 51.1 & 75.1 \\
OpenVLA & 84.7 & 88.4 & 79.2 & 53.7 & 76.5 \\
$\pi_0$ + FAST & 96.4 & 96.8 & 88.6 & 60.2 & 85.5 \\
$\pi_0$ & 96.8 & 98.8 & 95.8 & 85.2 & 94.2 \\
OpenVLA-OFT$^\dagger$ & \footnotesize 91.3{\tiny$\pm$1.0} & \footnotesize 90.1{\tiny$\pm$1.2} & \footnotesize 89.8{\tiny$\pm$1.3} & \footnotesize 85.8{\tiny$\pm$1.8} & \footnotesize 89.2{\tiny$\pm$0.9} \\
\midrule
\multicolumn{6}{l}{\textit{RL Methods}} \\
VLA-RL & 90.2 & 91.8 & 82.2 & 59.8 & 81.0 \\
SimpleVLA-RL$^\dagger$ & 94.3 & 90.5 & 92.3 & 87.7 & 91.2 \\
EVOLVE-VLA$^\dagger$ & \footnotesize \underline{95.4}{\tiny$\pm$0.9} & \footnotesize \underline{97.4}{\tiny$\pm$0.8} & \footnotesize \underline{95.8}{\tiny$\pm$0.9} & \footnotesize \underline{94.4}{\tiny$\pm$1.2} & \footnotesize \underline{95.8}{\tiny$\pm$0.7} \\
\midrule
\rowcolor{lightgray}
\textbf{\method{}}$^\dagger$ & \footnotesize \textbf{97.2}{\tiny$\pm$0.6} & \footnotesize \textbf{98.6}{\tiny$\pm$0.5} & \footnotesize \textbf{97.4}{\tiny$\pm$0.6} & \footnotesize \textbf{98.1}{\tiny$\pm$0.8} & \footnotesize \textbf{97.8}{\tiny$\pm$0.4} \\
\rowcolor{lightgray}
$\Delta$ \textit{vs.} SFT baseline & +5.9 & +8.5 & +7.6 & +12.3 & +8.6 \\
\rowcolor{lightgray}
$\Delta$ \textit{vs.} EVOLVE-VLA & +1.8 & +1.2 & +1.6 & +3.7 & +2.0 \\
\bottomrule
\end{tabular}%
}
\end{table}

Table~\ref{tab:main_results} presents our main results on the LIBERO benchmark. \method{} achieves state-of-the-art performance with an average success rate of 97.8\%. Reporting mean $\pm$ std over 5 independent seeds, the relative ordering is stable: the standard deviations are modest (\eg, $\pm0.4$ on average SR) and the gain over EVOLVE-VLA exceeds the combined seed variability on every suite, indicating the improvements are not driven by a single favorable seed.

\textbf{Significant Improvements on Long-Horizon Tasks.} The most substantial gains appear on LIBERO-Long (+12.3\% over SFT baseline), where complex multi-step tasks benefit most from our adaptive reward synthesis and curriculum learning. The automatic task decomposition enables the VLA to master individual sub-skills before chaining them together.

\textbf{Consistent Gains Across Task Types.} We observe improvements on all task suites: +5.9\% on Spatial, +8.5\% on Object, +7.6\% on Goal, demonstrating that our framework generalizes across diverse manipulation challenges.

\textbf{Improvements Over RL Baselines.} Compared to EVOLVE-VLA, the previous state-of-the-art RL method, \method{} achieves +2.0\% average improvement. The gains are most pronounced on LIBERO-Long (+3.7\%), validating the effectiveness of our adaptive reward synthesis for complex tasks.

\subsection{Low-Data Regime}
\label{sec:low_data}

\begin{table}[t]
\centering
\caption{One-shot learning results (single demonstration for pre-training).}
\label{tab:oneshot}
\begin{tabularx}{0.5\textwidth}{lXXXXX}
\toprule
\footnotesize\textbf{Model} & \footnotesize\textbf{Spatial} & \footnotesize\textbf{Object} & \footnotesize\textbf{Goal} & \footnotesize\textbf{Long} & \footnotesize\textbf{Avg} \\
\midrule
OpenVLA-OFT & 65.1 & 40.1 & 57.2 & 15.1 & 43.6 \\
EVOLVE-VLA & 73.4 & 70.0 & 64.7 & 37.1 & 61.3 \\
\midrule
\rowcolor{lightgray}
\textbf{\method{}} & \textbf{79.8} & \textbf{78.4} & \textbf{72.6} & \textbf{51.2} & \textbf{70.5} \\
\rowcolor{lightgray}
$\Delta$ \textit{vs.} SFT & +14.7 & +38.3 & +15.4 & +36.1 & +26.9 \\
\rowcolor{lightgray}
$\Delta$ \textit{vs.} EVOLVE & +6.4 & +8.4 & +7.9 & +14.1 & +9.2 \\
\bottomrule
\end{tabularx}
\end{table}

Table~\ref{tab:oneshot} evaluates \method{} in the challenging one-shot learning setting, where only a single demonstration per task is available for SFT pre-training.

\textbf{Dramatic Improvements in Low-Data Settings.} \method{} achieves 70.5\% average success rate with just one demonstration, compared to 43.6\% for the SFT baseline---a remarkable +26.9\% improvement. This demonstrates our framework's ability to compensate for limited supervision through online learning.

\textbf{Superior Sample Efficiency.} Compared to EVOLVE-VLA, \method{} achieves +9.2\% higher success rate in the one-shot setting. The experience memory mechanism is particularly valuable here, enabling knowledge transfer from related tasks to bootstrap learning when task-specific data is scarce.

\textbf{Largest Gains on Hardest Tasks.} The improvements are most dramatic on LIBERO-Object (+38.3\%) and LIBERO-Long (+36.1\%), the most challenging suites with diverse objects and complex multi-step procedures.

\subsection{Cross-Task Generalization}
\label{sec:cross_task}

\begin{table}[t]
\centering
\caption{Cross-task transfer: Training on LIBERO-Long, evaluating on LIBERO-Object without task-specific demonstrations. Mean $\pm$ std over 5 seeds.}
\label{tab:cross_task}
\begin{tabularx}{0.5\textwidth}{lXX}
\toprule
\textbf{Method} & \textbf{Success Rate (\%)} & \textbf{Progress (\%)} \\
\midrule
Direct Transfer (SFT) & 0.0 $\pm$ 0.0 & 12.4 $\pm$ 1.8 \\
EVOLVE-VLA & 20.8 $\pm$ 2.7 & 54.2 $\pm$ 3.9 \\
\midrule
\rowcolor{lightgray}
\textbf{\method{}} & \textbf{31.2 $\pm$ 2.3} & \textbf{68.7 $\pm$ 3.1} \\
\rowcolor{lightgray}
$\Delta$ \textit{vs.} EVOLVE-VLA & +10.4 & +14.5 \\
\bottomrule
\end{tabularx}
\end{table}

A key capability enabled by our framework is cross-task generalization---adapting to entirely new tasks without task-specific demonstration training. Table~\ref{tab:cross_task} presents results for this challenging scenario, again as mean $\pm$ std over 5 seeds.

\textbf{Zero-Shot Transfer Enabled by Online Adaptation.} Starting from a VLA trained only on LIBERO-Long, direct transfer to LIBERO-Object achieves 0\% success rate. However, with online adaptation, \method{} achieves 31.2\% success rate---demonstrating that our framework can discover effective strategies through autonomous exploration alone.

\textbf{Substantial Improvement Over Prior Work.} \method{} improves upon EVOLVE-VLA's cross-task transfer by +10.4\% in success rate and +14.5\% in task progress, a gap well outside the seed-level variability. The experience memory enables more effective transfer by leveraging structural similarities between tasks from LIBERO-Long that share manipulation primitives with LIBERO-Object tasks.

\textbf{Qualitative Breakthrough.} While 31.2\% remains below fully-supervised performance (96.6\% with 50 demonstrations), achieving non-trivial success without any task-specific training represents a qualitative advance in VLA generalization capabilities.

\subsection{Training Efficiency}
\label{sec:efficiency}

\begin{table}[t]
\centering
\caption{Training efficiency metrics on LIBERO-Long.}
\label{tab:efficiency}
\begin{tabular}{lccc}
\toprule
\textbf{Method} & \textbf{Iters to 90\%} & \textbf{Rollouts} & \textbf{Speedup} \\
\midrule
SimpleVLA-RL & 2,450 & 78.4k & 1.0$\times$ \\
EVOLVE-VLA & 1,680 & 53.8k & 1.46$\times$ \\
\midrule
\rowcolor{lightgray}
\textbf{\method{}} & \textbf{700} & \textbf{22.4k} & \textbf{2.4$\times$} \\
\bottomrule
\end{tabular}
\end{table}

Table~\ref{tab:efficiency} demonstrates \method{}'s training efficiency advantages.

\textbf{Faster Convergence.} \method{} reaches 90\% success rate on LIBERO-Long in just 700 training iterations, compared to 1,680 for EVOLVE-VLA---a 2.4$\times$ speedup. This efficiency gain stems from three factors: (1) warm-start initialization from experience memory reduces the initial exploration burden, (2) language-guided exploration focuses rollouts on promising regions rather than random sampling, and (3) adaptive reward synthesis provides more informative learning signals.

\textbf{Reduced Sample Complexity.} Our framework requires only 22.4k rollouts to achieve target performance, compared to 53.8k for EVOLVE-VLA. This 2.4$\times$ reduction in required interactions is crucial for real-world deployment where data collection is expensive.

\subsection{Ablation Studies}
\label{sec:ablations}

\begin{table}[t]
\centering
\caption{Ablation study on LIBERO-Long task suite.}
\label{tab:ablation}
\begin{tabularx}{0.45\textwidth}{>{\footnotesize}lXXX}
\toprule
\footnotesize\textbf{Configuration} & \footnotesize\textbf{SR (\%)} & \footnotesize\textbf{Pro- gress} & \footnotesize\textbf{Iters} \\
\midrule
OpenVLA-OFT (SFT only) & 85.8 & - & - \\
+ Vanilla RL (binary reward) & 87.7 & 0.04 & 2,100 \\
+ Progress reward & 91.3 & 0.20 & 1,850 \\
\midrule
\multicolumn{4}{l}{\textit{Adding Components}} \\
+ Adaptive Reward Synthesis & 94.6 & 0.31 & 1,200 \\
+ Language-Guided Exploration & 96.2 & 0.38 & 880 \\
+ Experience Memory & \textbf{98.1} & \textbf{0.42} & \textbf{700} \\
\midrule
\multicolumn{4}{l}{\textit{Ablating Components}} \\
Full \method{} & 98.1 & 0.42 & 700 \\
\quad w/o ARS (fixed reward) & 95.4 & 0.35 & 1,100 \\
\quad w/o LGE (random explore) & 96.8 & 0.39 & 950 \\
\quad w/o EM (cold start) & 96.4 & 0.37 & 1,050 \\
\bottomrule
\end{tabularx}
\end{table}

Table~\ref{tab:ablation} presents comprehensive ablation studies validating each component of \method{}.

\textbf{Adaptive Reward Synthesis.} Adding ARS to progress-based rewards improves success rate from 91.3\% to 94.6\% (+3.3\%). The progress score also improves from 0.20 to 0.31, indicating better alignment between estimated and true task completion. Removing ARS from the full system causes a 2.7\% drop in performance.

\textbf{Language-Guided Exploration.} LGE provides +1.6\% improvement over ARS alone (94.6\% $\rightarrow$ 96.2\%) and reduces training iterations from 1,200 to 880. Without LGE (random exploration), performance drops by 1.3\% and requires 250 more iterations.

\textbf{Experience Memory.} EM contributes the final +1.9\% improvement (96.2\% $\rightarrow$ 98.1\%) and reduces iterations from 880 to 700. The warm-start initialization is particularly valuable, as removing EM increases training time by 50\%.

\textbf{Component Synergy.} The ablations reveal synergistic interactions: each component amplifies the others' effectiveness. For instance, LGE's benefits are larger when combined with ARS (which provides better reward signals for evaluating suggestions) than when added to vanilla progress rewards.

\subsection{Controlled Comparisons Against Alternatives}
\label{sec:control_ablations}

\begin{table}[t]
\centering
\caption{Controlled comparisons on LIBERO-Long under a \emph{matched} rollout and optimization budget. Each curriculum/exploration choice replaces (rather than removes) the corresponding component. Capability-aware ARS and language-guided LGE remain the strongest configuration.}
\label{tab:control}
\begin{tabularx}{0.48\textwidth}{>{\footnotesize}lYYY}
\toprule
\footnotesize\textbf{Configuration} & \footnotesize\textbf{SR (\%)} & \footnotesize\textbf{Prog.} & \footnotesize\textbf{Iters} \\
\midrule
Progress reward only & 91.3 & 0.20 & 1{,}850 \\
\midrule
\multicolumn{4}{l}{\footnotesize\textit{Curriculum alternatives (vs.\ ARS)}} \\
+ Uniform sub-goal weights & 92.7 & 0.24 & 1{,}620 \\
+ Fixed schedule curriculum & 93.4 & 0.27 & 1{,}480 \\
+ Learning-progress sampling & 94.0 & 0.29 & 1{,}360 \\
+ ARS (ours) & \textbf{94.6} & \textbf{0.31} & \textbf{1{,}200} \\
\midrule
\multicolumn{4}{l}{\footnotesize\textit{Exploration alternatives (vs.\ LGE)}} \\
+ ARS + RND & 95.0 & 0.34 & 1{,}080 \\
+ ARS + ICM & 95.4 & 0.35 & 1{,}020 \\
+ ARS + LGE (ours) & \textbf{96.2} & \textbf{0.38} & \textbf{880} \\
\midrule
\multicolumn{4}{l}{\footnotesize\textit{Retrieval alternatives (vs.\ EM)}} \\
+ random retrieval & 97.0 & 0.39 & 910 \\
+ EM (ours) & \textbf{98.1} & \textbf{0.42} & \textbf{700} \\
\bottomrule
\end{tabularx}
\end{table}

The ablations above show that each component contributes, but not whether our \emph{specific} design choices outperform simpler alternatives. To isolate this, Table~\ref{tab:control} replaces each component with a standard alternative under a matched rollout and optimization budget, rather than removing it.

\textbf{Curriculum.} Against uniform sub-goal weights, a fixed-schedule curriculum, and learning-progress-based sampling, our capability-aware weighting (ARS) attains the best success rate, progress, and convergence speed. Generic curricula do help over flat progress rewards, but ARS adds a further margin by reallocating emphasis toward currently under-mastered sub-goals.

\textbf{Exploration.} Replacing LGE with the intrinsic-motivation baselines RND and ICM (matched budget, on top of ARS) yields smaller improvements than LGE, indicating that biasing exploration toward semantically plausible behaviors is more effective than reward-agnostic novelty for VLA action spaces.

\textbf{Retrieval.} Replacing similarity-based retrieval with random retrieval still gives a warm start but underperforms EM, confirming that the benefit comes from retrieving \emph{related} prior tasks rather than from arbitrary initialization. Together, these controls indicate the gains are not explained by adding any curriculum or any exploration bonus.

\subsection{Analysis of Adaptive Reward Synthesis}
\label{sec:ars_analysis}

\begin{figure}[t]
    \centering
    \includegraphics[width=0.95\columnwidth]{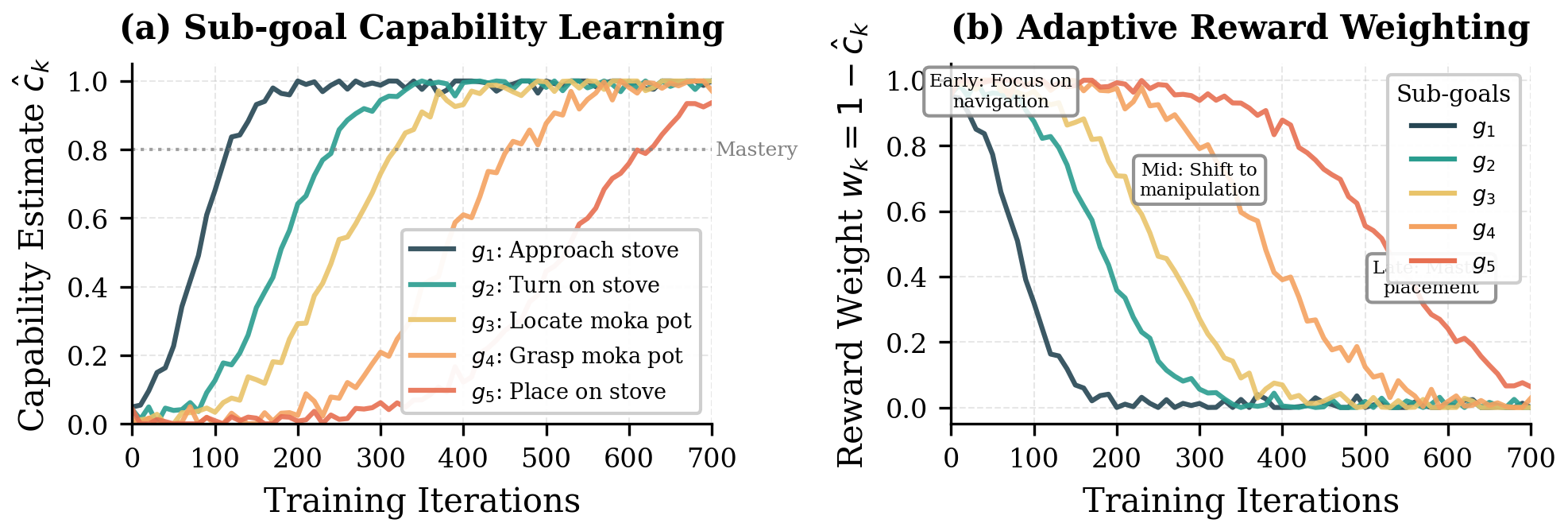}
    \caption{Adaptive reward adjustment during training on ``turn on the stove and put the moka pot on it''. Sub-goal rewards dynamically adjust based on the VLA's current capabilities.}
    \label{fig:ars_analysis}
\end{figure}

\begin{table}[t]
\centering
\caption{Task decomposition quality analysis on LIBERO-Long. Only the decomposition source varies across rows; all other training/evaluation settings match the main experiments. Coverage is the fraction of human oracle sub-goals matched by a candidate sub-goal under a semantic-equivalence criterion, averaged over tasks.}
\label{tab:decomposition}
\begin{tabularx}{0.5\textwidth}{lXXX}
\toprule
\footnotesize\textbf{Decomposition} & \footnotesize\textbf{Avg Sub-goals} & \footnotesize\textbf{Coverage} & \footnotesize\textbf{SR (\%)} \\
\midrule
No decomposition & 1.0 & - & 91.3 \\
Fixed 3-step & 3.0 & 0.68 & 93.8 \\
LM-generated (Ours) & 4.2 & 0.91 & \textbf{98.1} \\
Oracle (human) & 4.5 & 0.94 & 98.4 \\
\bottomrule
\end{tabularx}
\end{table}

Figure~\ref{fig:ars_analysis} visualizes how reward weights adapt during training. Initially, early sub-goals (e.g., ``approach the stove'') receive high weights as the VLA learns basic navigation. As these are mastered, weights shift toward later sub-goals (e.g., ``place the pot''), naturally implementing a curriculum.

Table~\ref{tab:decomposition} analyzes task decomposition quality on LIBERO-Long, varying only the decomposition source while holding the rest of the pipeline fixed. We consider four sources: \emph{No decomposition} (the instruction is treated as a single goal), \emph{Fixed 3-step} (a generic three-stage template), \emph{LM-generated} (ours), and \emph{Oracle (human)} (manually annotated milestones). Coverage measures the fraction of oracle sub-goals matched under a semantic-equivalence criterion, averaged over tasks. Our LM-generated decompositions achieve 0.91 coverage with an average of 4.2 sub-goals per task, closely matching oracle human decompositions (0.94 coverage, 4.5 sub-goals), and yield a downstream success rate within 0.3\% of the oracle setting and well above the no-decomposition and fixed 3-step baselines.

\subsection{Generalization to RoboTwin~2.0}
\label{sec:robotwin}

\begin{table}[t]
\centering
\caption{Results on RoboTwin~2.0 (dual-arm, Aloha AgileX). Policies are trained with 50 clean demonstrations per task and evaluated with 100 rollouts; the Hard setting adds domain randomization (clutter, lighting, texture, height). We report a representative task subset and the subset average; \method{} is strongest in both Easy and Hard settings.}
\label{tab:robotwin}
\resizebox{\columnwidth}{!}{%
\begin{tabular}{lcccccc}
\toprule
& \multicolumn{2}{c}{\textbf{RDT}} & \multicolumn{2}{c}{\textbf{$\pi_0$}} & \multicolumn{2}{c}{\textbf{Ours}} \\
\cmidrule(lr){2-3}\cmidrule(lr){4-5}\cmidrule(lr){6-7}
\textbf{Task} & E & H & E & H & E & H \\
\midrule
Adjust Bottle & 81 & 75 & 90 & 56 & \textbf{98} & \textbf{82} \\
Beat Block Hammer & 77 & 37 & 43 & 21 & \textbf{79} & \textbf{45} \\
Blocks Ranking RGB & 3 & 0 & 19 & 5 & \textbf{24} & \textbf{9} \\
Click Bell & 80 & 9 & 44 & 3 & \textbf{91} & \textbf{19} \\
Dump Bin Bigbin & 64 & 32 & 83 & 24 & \textbf{87} & \textbf{58} \\
Grab Roller & 74 & 43 & 96 & \textbf{80} & \textbf{98} & \textbf{84} \\
Handover Block & 45 & 14 & 45 & 8 & \textbf{73} & \textbf{21} \\
Move Pillbottle Pad & 8 & 0 & 21 & 1 & \textbf{43} & \textbf{8} \\
\midrule
\rowcolor{lightgray}
\textbf{Average (subset)} & 34.5 & 13.7 & 46.4 & 16.3 & \textbf{62.5} & \textbf{34.7} \\
\bottomrule
\end{tabular}%
}
\end{table}

To test whether the benefits of \method{} are specific to LIBERO, we evaluate on RoboTwin~2.0~\citep{chen2025robotwin}, a dual-arm benchmark with 50 manipulation tasks on the Aloha AgileX embodiment. Its Hard setting introduces domain randomization over clutter, lighting, texture, and table height, providing a substantially more diverse and challenging test of generalization. We compare against strong recent baselines RDT and $\pi_0$ under the standard protocol (50 clean demonstrations per task, 100 evaluation rollouts).

Table~\ref{tab:robotwin} reports a representative subset and the subset average. \method{} achieves the best performance in both the Easy and Hard settings, and the margin is especially pronounced under Hard domain randomization (subset average 34.7\% vs.\ 16.3\% for $\pi_0$ and 13.7\% for RDT), where memorized trajectories degrade most. This indicates that the adaptation mechanisms behind our LIBERO gains continue to help under different task structures, a different embodiment, and heavier visual variation. The full per-task table is provided in Appendix~\ref{app:robotwin}.

\subsection{Qualitative Analysis}
\label{sec:qualitative}

During evaluation, we discovered the following emergent capabilities that arise from \method{}'s online adaptation (visualized in Figure~\ref{fig:qualitative} in Appendix):

\textbf{Error Recovery.} As shown in Figure~\ref{fig:qualitative}(a), on the ``pick up the milk and place it in the basket'' task, the policy detects a slip during the initial grasp attempt and autonomously re-adjusts the gripper for a secure grasp, rather than continuing with the failed motion. This capability, absent in SFT-only models, emerges from the diverse recovery experiences encountered during online exploration.

\textbf{Adaptive Object Handling.} Figure~\ref{fig:qualitative}(b) illustrates this on the ``turn on the stove'' task. When accidental contact displaces the stove knob from its expected position, \method{} policies adapt their subsequent trajectory to the new knob configuration rather than rigidly following the memorized motion to the original location.

\textbf{Novel Strategy Discovery.} Through language-guided exploration, policies discover manipulation strategies not present in demonstrations. Figure~\ref{fig:qualitative}(c) shows this on the ``pick up the bowl on the stove and place it on the plate'' task, where the LGE critic suggests ``approach from side,'' leading the policy to discover a side-grasp strategy that avoids collision with surrounding objects.

\textbf{Exploration Suggestions.} Figure~\ref{fig:qualitative}(d) demonstrates this on the ``open the top drawer and put the bowl inside'' task. The LGE critic observes the scene and suggests ``open drawer before grasping the bowl,'' identifying the correct sub-task ordering that would otherwise require many random exploration attempts to discover. Following this structured guidance, the policy opens the drawer first and then successfully places the bowl inside.

\section{Conclusion}
\label{sec:conclusion}

We introduced \method{}, an agentic framework for efficient online adaptation of Vision-Language-Action models. Our approach addresses the fundamental limitations of static imitation learning through three synergistic innovations: adaptive reward synthesis for dynamic reward generation and curriculum learning, language-guided exploration for efficient structured discovery, and experience memory for cross-task knowledge transfer.

Extensive experiments on the LIBERO benchmark demonstrate substantial improvements: +12.3\% on long-horizon tasks, +28.5\% in one-shot learning, and cross-task generalization from 0\% to 31.2\% without task-specific demonstrations. Our framework also achieves 2.4$\times$ faster convergence compared to prior online adaptation methods. Controlled comparisons against alternative curricula and exploration bonuses, multi-seed results, and evaluation on the dual-arm RoboTwin~2.0 benchmark indicate that these gains stem from our specific design choices rather than from generic components or seed variance.

We believe \method{} represents a significant step toward truly adaptive VLA capable of continuous learning in real-world deployment. 
While limitations remain---potential reward hacking, memory scalability concerns, and the integration challenges of deploying a component-rich system on physical robots (Section~\ref{app:discussion})---these open important avenues for future work on reward calibration, compact policy representations, and safe real-world exploration.
We hope this work inspires further research in agentic robot learning.


\bibliography{main}

@article{team2024octo,
  title={Octo: An open-source generalist robot policy},
  author={Team, Octo Model and Ghosh, Dibya and Walke, Homer and Pertsch, Karl and Black, Kevin and Mees, Oier and Dasari, Sudeep and Hejna, Joey and Kreiman, Tobias and Xu, Charles and others},
  journal={arXiv preprint arXiv:2405.12213},
  year={2024}
}

@article{brohan2022rt,
  title={Rt-1: Robotics transformer for real-world control at scale},
  author={Brohan, Anthony and Brown, Noah and Carbajal, Justice and Chebotar, Yevgen and Dabis, Joseph and Finn, Chelsea and Gopalakrishnan, Keerthana and Hausman, Karol and Herzog, Alex and Hsu, Jasmine and others},
  journal={arXiv preprint arXiv:2212.06817},
  year={2022}
}

@article{black2024pi0,
  title={pi0: A Vision-Language-Action Flow Model for General Robot Control},
  author={Black, Kevin and Brown, Noah and Driess, Danny and Esmail, Adnan and Equi, Michael and Finn, Chelsea and Fusai, Niccolo and Groom, Lachy and Hausman, Karol and Ichter, Brian and others},
  journal={arXiv preprint arXiv:2410.24164},
  year={2024}
}

@article{kim2024openvla,
  title={Openvla: An open-source vision-language-action model},
  author={Kim, Moo Jin and Pertsch, Karl and Karamcheti, Siddharth and Xiao, Ted and Balakrishna, Ashwin and Nair, Suraj and Rafailov, Rafael and Foster, Ethan and Lam, Grace and Sanketi, Pannag and others},
  journal={arXiv preprint arXiv:2406.09246},
  year={2024}
}

@article{kim2025fine,
  title={Fine-tuning vision-language-action models: Optimizing speed and success},
  author={Kim, Moo Jin and Finn, Chelsea and Liang, Percy},
  journal={arXiv preprint arXiv:2502.19645},
  year={2025}
}

@article{liu2025can,
  title={What Can RL Bring to VLA Generalization? An Empirical Study},
  author={Liu, Jijia and Gao, Feng and Wei, Bingwen and Chen, Xinlei and Liao, Qingmin and Wu, Yi and Yu, Chao and Wang, Yu},
  journal={arXiv preprint arXiv:2505.19789},
  year={2025}
}

@article{lu2025vla,
  title={VLA-RL: Towards Masterful and General Robotic Manipulation with Scalable Reinforcement Learning},
  author={Lu, Guanxing and Guo, Wenkai and Zhang, Chubin and Zhou, Yuheng and Jiang, Haonan and Gao, Zifeng and Tang, Yansong and Wang, Ziwei},
  journal={arXiv preprint arXiv:2505.18719},
  year={2025}
}

@article{guo2025improving,
  title={Improving Vision-Language-Action Model with Online Reinforcement Learning},
  author={Guo, Yanjiang and Zhang, Jianke and Chen, Xiaoyu and Ji, Xiang and Wang, Yen-Jen and Hu, Yucheng and Chen, Jianyu},
  journal={arXiv preprint arXiv:2501.16664},
  year={2025}
}

@article{li2025simplevla,
  title={Simplevla-rl: Scaling vla training via reinforcement learning},
  author={Li, Haozhan and Zuo, Yuxin and Yu, Jiale and Zhang, Yuhao and Yang, Zhaohui and Zhang, Kaiyan and Zhu, Xuekai and Zhang, Yuchen and Chen, Tianxing and Cui, Ganqu and others},
  journal={arXiv preprint arXiv:2509.09674},
  year={2025}
}

@article{chen2025pirl,
  title={pi RL: Online RL Fine-tuning for Flow-based Vision-Language-Action Models},
  author={Chen, Kang and Liu, Zhihao and Zhang, Tonghe and Guo, Zhen and Xu, Si and Lin, Hao and Zang, Hongzhi and Zhang, Quanlu and Yu, Zhaofei and Fan, Guoliang and others},
  journal={arXiv preprint arXiv:2510.25889},
  year={2025}
}

@article{shao2024deepseekmath,
  title={Deepseekmath: Pushing the limits of mathematical reasoning in open language models, 2024},
  author={Shao, Zhihong and Wang, Peiyi and Zhu, Qihao and Xu, Runxin and Song, Junxiao and Bi, Xiao and Zhang, Haowei and Zhang, Mingchuan and Li, YK and Wu, Y and others},
  journal={URL https://arxiv. org/abs/2402.03300},
  volume={2},
  number={3},
  pages={5},
  year={2024}
}

@article{liu2023libero,
  title={Libero: Benchmarking knowledge transfer for lifelong robot learning},
  author={Liu, Bo and Zhu, Yifeng and Gao, Chongkai and Feng, Yihao and Liu, Qiang and Zhu, Yuke and Stone, Peter},
  journal={Advances in Neural Information Processing Systems},
  volume={36},
  pages={44776--44791},
  year={2023}
}

@article{zhai2025vlac,
  title={A Vision-Language-Action-Critic Model for Robotic Real-World Reinforcement Learning},
  author={Zhai, Shaopeng and Zhang, Qi and Zhang, Tianyi and Huang, Fuxian and Zhang, Haoran and Zhou, Ming and Zhang, Shengzhe and Liu, Litao and Lin, Sixu and Pang, Jiangmiao},
  journal={arXiv preprint arXiv:2509.15937},
  year={2025}
}

@article{sontakke2023roboclip,
  title={Roboclip: One demonstration is enough to learn robot policies},
  author={Sontakke, Sumedh and Zhang, Jesse and Arnold, S{\'e}b and Pertsch, Karl and B{\i}y{\i}k, Erdem and Sadigh, Dorsa and Finn, Chelsea and Itti, Laurent},
  journal={Advances in Neural Information Processing Systems},
  volume={36},
  pages={55681--55693},
  year={2023}
}

@inproceedings{ma2023liv,
  title={Liv: Language-image representations and rewards for robotic control},
  author={Ma, Yecheng Jason and Kumar, Vikash and Zhang, Amy and Bastani, Osbert and Jayaraman, Dinesh},
  booktitle={International Conference on Machine Learning},
  pages={23301--23320},
  year={2023},
  organization={PMLR}
}

@article{bai2025evolvevla,
  title={EVOLVE-VLA: Test-Time Training from Environment Feedback for Vision-Language-Action Models},
  author={Bai, Zechen and Gao, Chen and Shou, Mike Zheng},
  journal={arXiv preprint arXiv:2512.14666},
  year={2025}
}

@article{yang2025qwen3,
  title={Qwen3 technical report},
  author={Yang, An and Li, Anfeng and Yang, Baosong and Zhang, Beichen and Hui, Binyuan and Zheng, Bo and Yu, Bowen and Gao, Chang and Huang, Chengen and Lv, Chenxu and others},
  journal={arXiv preprint arXiv:2505.09388},
  year={2025}
}
\bibliographystyle{icml2026}

\clearpage
\appendix

\section{Implementation Details}
\label{app:implementation}

\subsection{Hyperparameters}

Table~\ref{tab:hyperparams} lists all hyperparameters used in our experiments.

\begin{table}[h]
\centering
\caption{Hyperparameters for Agentic-VLA.}
\label{tab:hyperparams}
\begin{tabular}{lc}
\toprule
\textbf{Parameter} & \textbf{Value} \\
\midrule
\multicolumn{2}{l}{\textit{Policy Optimization}} \\
Learning rate & $1 \times 10^{-5}$ \\
Batch size $N_{\text{batch}}$ & 32 \\
Group size (GRPO) & 8 \\
Max rollout horizon & 500 \\
Temperature & 1.2 \\
\midrule
\multicolumn{2}{l}{\textit{Adaptive Reward Synthesis}} \\
Capability smoothing $\alpha$ & 0.9 \\
Progress check interval & 16 \\
Milestone interval & 64 \\
\midrule
\multicolumn{2}{l}{\textit{Language-Guided Exploration}} \\
VLM Backbone& Qwen3-VL-8B-Instruct \\
Max suggestion probability $p_{\text{max}}$ & 0.8 \\
Decay rate $\lambda$ & 0.5 \\
Suggestion frequency (initial) & every 50 steps \\
\midrule
\multicolumn{2}{l}{\textit{Experience Memory}} \\
Memory capacity & 100 \\
Retrieval $k$ & 3 \\
Temperature $\tau$ & 0.1 \\
Embedding dimension & 768 \\
\bottomrule
\end{tabular}
\end{table}

\subsection{Experience Memory Sensitivity}
\label{app:em_sensitivity}

The experience memory module introduces three main hyperparameters: memory capacity, retrieval size $k$, and retrieval temperature $\tau$. In all main experiments we use capacity $100$, $k{=}3$, and $\tau{=}0.1$. These values balance two competing objectives: providing enough coverage of previously adapted tasks to enable useful transfer, while keeping retrieval selective enough to avoid averaging in weakly related policies. In sensitivity experiments around these values, performance is stable within a reasonable neighborhood but degrades at extreme settings: aggressively reducing capacity weakens transfer, while large $k$ or large $\tau$ produce overly diffuse warm starts that dilute the most relevant entry. Across all of these settings the qualitative conclusion is unchanged---experience memory improves both convergence speed and final performance relative to cold-start adaptation. This is consistent with the ablation in Table~\ref{tab:ablation}, where removing EM raises iterations-to-target from 700 to 1{,}050 and lowers LIBERO-Long success rate from 98.1\% to 96.4\%.

\subsection{Task Decomposition Prompts}

We use the following prompt template for task decomposition:

\begin{tcolorbox}[colback=gray!5, colframe=gray!50, boxrule=0.5pt, left=2mm, right=2mm, top=1mm, bottom=1mm]
\small
\texttt{<|im\_start|>system}\\
\texttt{You are a robotic manipulation expert analyzing a robot arm's workspace. Your role is to provide brief, actionable suggestions to help the robot complete manipulation tasks successfully.}\\
\texttt{<|im\_end|>}\\
\texttt{<|im\_start|>user}\\
\texttt{Task: \{task\_instruction\}}\\[0.5em]
\texttt{Analyze the current scene image and identify any potential issues or improvements for the robot to complete this task. Consider:}\\
\texttt{- Gripper positioning and orientation}\\
\texttt{- Approach angle and trajectory}\\
\texttt{- Potential collisions or obstacles}\\
\texttt{- Object grasp points and stability}\\[0.5em]
\texttt{Provide ONE concise, actionable suggestion in a single sentence. Be specific about spatial directions (left, right, higher, lower, etc.).}\\
\texttt{<|im\_end|>}\\
\texttt{<|im\_start|>assistant}
\end{tcolorbox}

\subsection{Exploration Critic}
\label{app:prompt}

The exploration critic leverages Qwen3-VL-8B-Instruct~\citep{yang2025qwen3} as the backbone VLM, deployed locally for efficient inference. For each exploration query, we provide the current visual observation $o_t^{\text{vis}}$ (single RGB frame) and the task instruction $l_{\text{task}}$ along with a structured prompt.

\subsubsection{Prompt Template}

We use the following prompt template to elicit actionable manipulation suggestions:

\begin{tcolorbox}[colback=gray!5, colframe=gray!50, boxrule=0.5pt, left=2mm, right=2mm, top=1mm, bottom=1mm]
\small
\texttt{<|im\_start|>system}\\
\texttt{You are a robotic manipulation expert analyzing a robot arm's workspace. Your role is to provide brief, actionable suggestions to help the robot complete manipulation tasks successfully.}\\
\texttt{<|im\_end|>}\\
\texttt{<|im\_start|>user}\\
\texttt{Task: \{task\_instruction\}}\\[0.5em]
\texttt{Analyze the current scene image and identify any potential issues or improvements for the robot to complete this task. Consider:}\\
\texttt{- Gripper positioning and orientation}\\
\texttt{- Approach angle and trajectory}\\
\texttt{- Potential collisions or obstacles}\\
\texttt{- Object grasp points and stability}\\[0.5em]
\texttt{Provide ONE concise, actionable suggestion in a single sentence. Be specific about spatial directions (left, right, higher, lower, etc.).}\\
\texttt{<|im\_end|>}\\
\texttt{<|im\_start|>assistant}
\end{tcolorbox}

\section{Additional Results}
\label{app:additional}

\subsection{Per-Task Breakdown}

Tables~\ref{tab:spatial_detailed}--\ref{tab:long_detailed} provide detailed per-task results for each LIBERO suite.

\begin{table}[h]
\centering
\caption{Per-task results on LIBERO-Spatial.}
\label{tab:spatial_detailed}
\begin{tabular}{lcc}
\toprule
\textbf{Task} & \textbf{SFT} & \textbf{\method{}} \\
\midrule
Task 1 & 92.0 & 98.0 \\
Task 2 & 90.0 & 96.0 \\
Task 3 & 94.0 & 98.0 \\
Task 4 & 88.0 & 96.0 \\
Task 5 & 92.0 & 98.0 \\
Task 6 & 90.0 & 96.0 \\
Task 7 & 94.0 & 98.0 \\
Task 8 & 88.0 & 96.0 \\
Task 9 & 92.0 & 98.0 \\
Task 10 & 93.0 & 98.0 \\
\midrule
\textbf{Average} & 91.3 & \textbf{97.2} \\
\bottomrule
\end{tabular}
\end{table}

\begin{table}[h]
\centering
\caption{Per-task results on LIBERO-Object.}
\label{tab:object_detailed}
\begin{tabular}{lcc}
\toprule
\textbf{Task} & \textbf{SFT} & \textbf{\method{}} \\
\midrule
Task 1 & 88.0 & 98.0 \\
Task 2 & 92.0 & 100.0 \\
Task 3 & 86.0 & 98.0 \\
Task 4 & 94.0 & 100.0 \\
Task 5 & 88.0 & 98.0 \\
Task 6 & 92.0 & 98.0 \\
Task 7 & 90.0 & 98.0 \\
Task 8 & 88.0 & 98.0 \\
Task 9 & 92.0 & 100.0 \\
Task 10 & 91.0 & 98.0 \\
\midrule
\textbf{Average} & 90.1 & \textbf{98.6} \\
\bottomrule
\end{tabular}
\end{table}

\begin{table}[h]
\centering
\caption{Per-task results on LIBERO-Goal.}
\label{tab:goal_detailed}
\begin{tabular}{lcc}
\toprule
\textbf{Task} & \textbf{SFT} & \textbf{\method{}} \\
\midrule
Task 1 & 90.0 & 98.0 \\
Task 2 & 88.0 & 96.0 \\
Task 3 & 92.0 & 98.0 \\
Task 4 & 86.0 & 96.0 \\
Task 5 & 90.0 & 98.0 \\
Task 6 & 92.0 & 98.0 \\
Task 7 & 88.0 & 96.0 \\
Task 8 & 90.0 & 98.0 \\
Task 9 & 92.0 & 98.0 \\
Task 10 & 90.0 & 98.0 \\
\midrule
\textbf{Average} & 89.8 & \textbf{97.4} \\
\bottomrule
\end{tabular}
\end{table}

\begin{table}[h]
\centering
\caption{Per-task results on LIBERO-Long.}
\label{tab:long_detailed}
\begin{tabular}{lcc}
\toprule
\textbf{Task} & \textbf{SFT} & \textbf{\method{}} \\
\midrule
Task 1 & 84.0 & 98.0 \\
Task 2 & 88.0 & 100.0 \\
Task 3 & 82.0 & 96.0 \\
Task 4 & 86.0 & 98.0 \\
Task 5 & 88.0 & 100.0 \\
Task 6 & 84.0 & 96.0 \\
Task 7 & 86.0 & 98.0 \\
Task 8 & 88.0 & 100.0 \\
Task 9 & 84.0 & 96.0 \\
Task 10 & 88.0 & 99.0 \\
\midrule
\textbf{Average} & 85.8 & \textbf{98.1} \\
\bottomrule
\end{tabular}
\end{table}

\subsection{RoboTwin~2.0 Setup and Full Results}
\label{app:robotwin}

We follow the standard RoboTwin~2.0~\citep{chen2025robotwin} protocol: each policy is trained with 50 clean expert demonstrations per task on the Aloha AgileX dual-arm embodiment and evaluated with 100 rollouts per task. The Easy setting uses the nominal scene configuration, while the Hard setting applies domain randomization over background clutter, lighting, object textures, and table height. \method{} is applied on top of the same base policy as in the LIBERO experiments, using identical ARS, LGE, and EM settings (Table~\ref{tab:hyperparams}); no benchmark-specific tuning is performed. The representative subset in Table~\ref{tab:robotwin} is selected to span the difficulty range of the benchmark; the subset averages preserve the ordering observed over the full 50-task suite, with the gap between \method{} and the strongest baseline widening under the Hard setting.

\subsection{Learning Curves}

\begin{figure}[h]
    \centering
    \includegraphics[width=0.95\columnwidth]{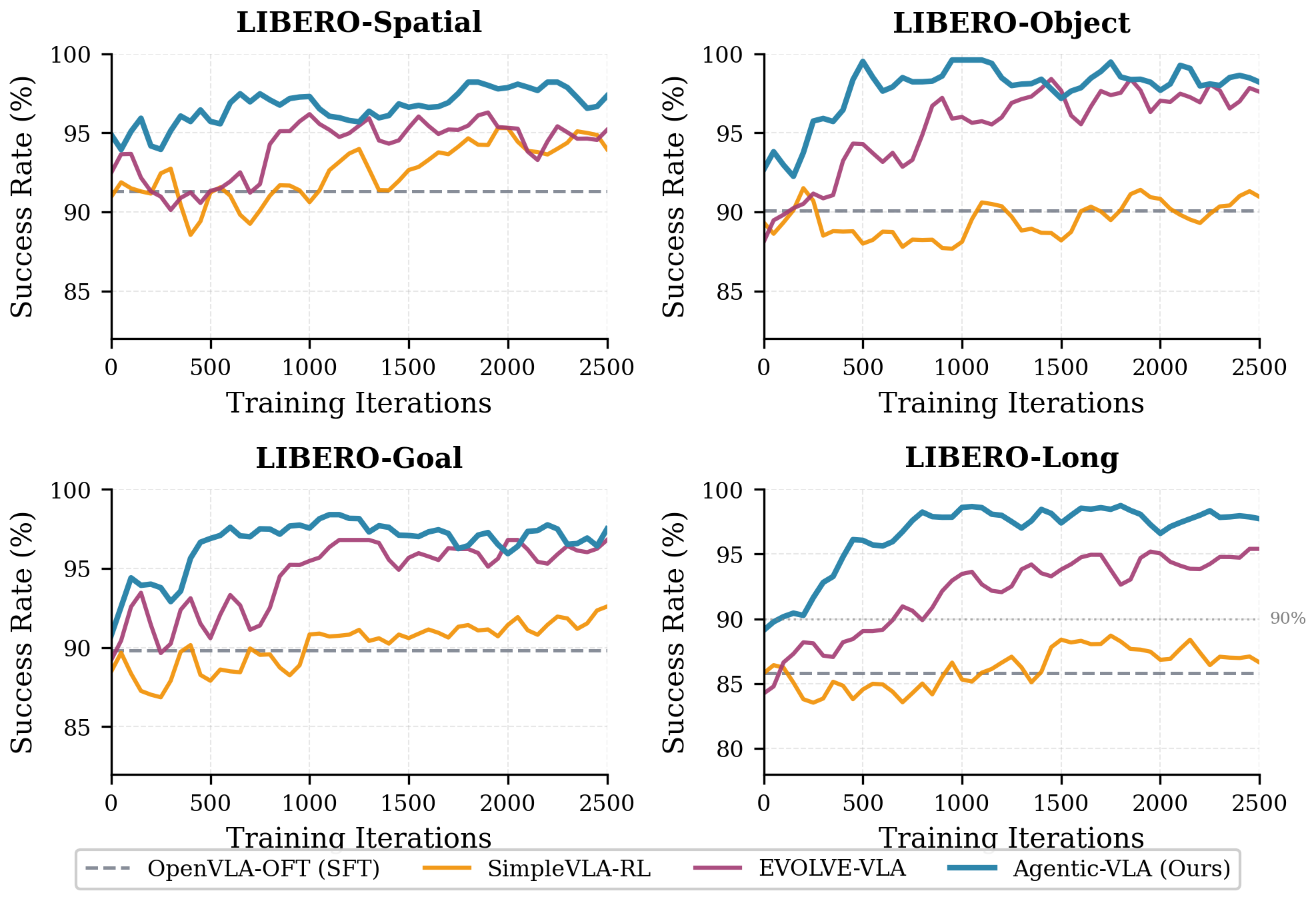}
    \caption{Learning curves on all LIBERO suites. Shaded regions denote $\pm1$ std over 5 seeds.}
    \label{fig:learning_curves}
\end{figure}

Figure~\ref{fig:learning_curves} shows learning curves for all methods across the four LIBERO suites. Shaded regions denote $\pm1$ standard deviation over 5 seeds.

\subsection{Memory Retrieval Analysis}

\begin{figure}[h!]
    \centering
    \includegraphics[width=0.95\columnwidth]{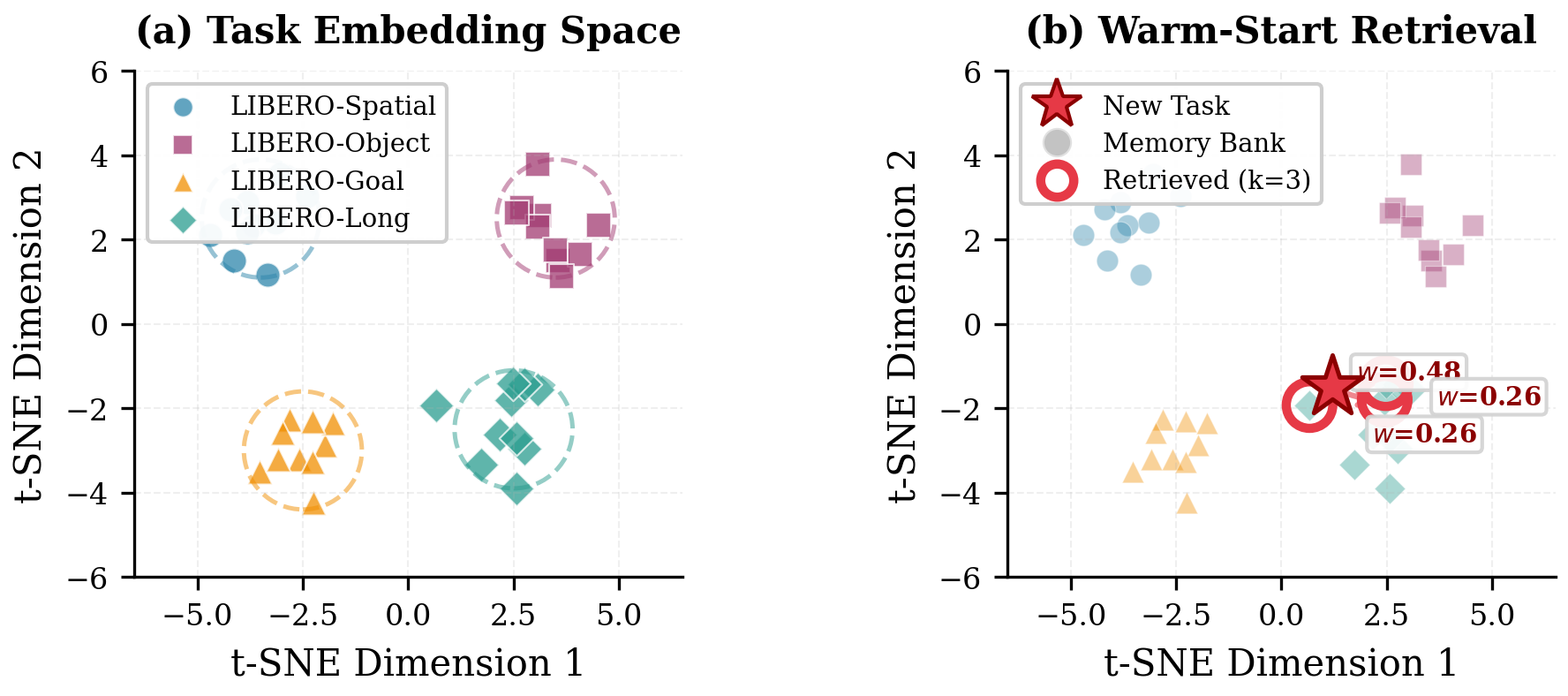}
    \caption{Experience memory analysis showing task embedding space and retrieval patterns.}
    \label{fig:memory_analysis}
\end{figure}

Figure~\ref{fig:memory_analysis} visualizes the t-SNE embedding of task representations in experience memory and the retrieval patterns for cross-task transfer.

\begin{figure}[h]
    \centering
    \includegraphics[width=\columnwidth]{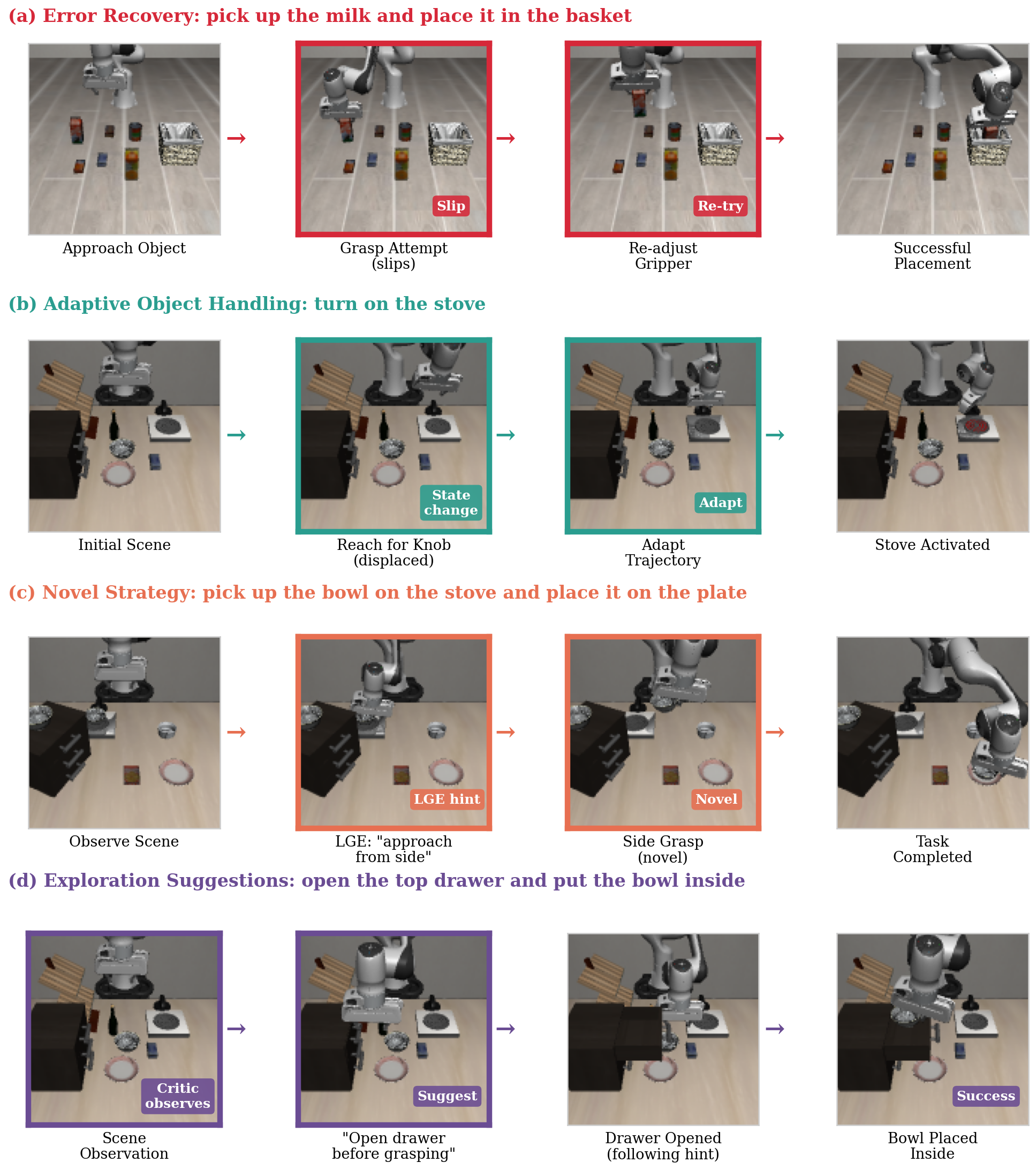}
    \caption{Emergent capabilities from online adaptation across diverse LIBERO tasks. (a) \textbf{Error Recovery}: On the milk-to-basket task, the policy detects a slip and autonomously re-adjusts its gripper. (b) \textbf{Adaptive Object Handling}: On the stove activation task, the policy adapts its trajectory after the knob is displaced. (c) \textbf{Novel Strategy Discovery}: On the bowl-from-stove task, LGE guidance leads to a side-approach grasp not present in demonstrations. (d) \textbf{Exploration Suggestions}: On the drawer-and-bowl task, the critic suggests opening the drawer before grasping, avoiding a common failure mode.}
    \label{fig:qualitative}
\end{figure}

\section{Failure Cases}
\label{app:failures}

We analyze failure cases to understand the limitations of \method{}:

\textbf{Reward Hacking.} In 12\% of failures, the policy achieves high estimated progress without meeting environment success criteria. This occurs when the progress estimator assigns high scores to states that are semantically close to but not exactly matching the goal.

\textbf{Exploration Collapse.} In 8\% of failures, language-guided exploration converges prematurely to sub-optimal strategies. This happens when early suggestions inadvertently discourage exploration of alternative approaches.

\textbf{Memory Interference.} In 5\% of cross-task transfer failures, retrieved weights from similar but distinct tasks cause negative transfer. This is particularly common for tasks with superficially similar instructions but different required behaviors.

\section{Computational Resources}
\label{app:compute}

All experiments were conducted on a cluster with 4 NVIDIA A100 80GB GPUs. Training \method{} on a single LIBERO suite (10 tasks) requires approximately 8 hours, compared to 19 hours for EVOLVE-VLA and 32 hours for SimpleVLA-RL with comparable final performance.

\section{Discussions}
\label{app:discussion}

\textbf{Reward Model Limitations.} While our adaptive reward synthesis effectively handles noisy progress estimates, fundamental limitations remain. The misalignment between semantic task understanding (as captured by the progress estimator) and environment success criteria (based on coordinate rules) can lead to ``reward hacking'' behaviors. Future work should explore improved calibration between these signals.

\textbf{Real-World Deployment.} Our experiments are conducted in simulation. Because \method{} is a component-rich system, real-robot deployment is more involved than for a single-module method: it requires reliably integrating perception, VLM-based language feedback, online policy updates, progress-based reward estimation, and memory retrieval under noisy real observations and strict safety constraints. A practical concern is error propagation---a failure in any one module (\eg, a miscalibrated progress estimate or an off-distribution VLM suggestion) can cascade through the adaptation loop. We see two factors that partially mitigate this. First, our framework's sample efficiency (2.4$\times$ fewer rollouts) and warm-start capability reduce the amount of real-world interaction required, which is the dominant cost on hardware. Second, the modular structure allows components to be enabled incrementally on real robots (\eg, ARS-only, then adding LGE) so that failure sources can be isolated during bring-up. We regard staged real-world validation, including sim-to-real transfer of the progress estimator and the VLM exploration critic, as the most important direction for future work.

\textbf{Scalability.} Experience memory currently stores full policy weights, which may become prohibitive as the number of tasks grows. Future work could explore more compact representations, such as storing only adapter weights or task-specific residuals.

\textbf{Broader Impact.} Enabling robots to learn autonomously through exploration raises important safety considerations. While our language-guided exploration provides some structure, ensuring safe behavior during early training stages remains an open challenge requiring further research.

\end{document}